\pdfoutput=1

\documentclass[11pt]{article}

\usepackage[final]{acl}

\usepackage{times}
\usepackage{latexsym}

\usepackage[T1]{fontenc}

\usepackage[utf8]{inputenc}
\usepackage{inconsolata}
\usepackage{microtype}

\usepackage{inconsolata}
\usepackage{hyperref}
\usepackage{url}
\usepackage{hyperref}       
\usepackage{url}            
\usepackage{booktabs}       
\usepackage{amsfonts}       
\usepackage{nicefrac}       
\usepackage{microtype}      
\usepackage{xcolor}         
\usepackage{url}            
\usepackage{booktabs}       
\usepackage{amsfonts}       
\usepackage{nicefrac}       
\usepackage{microtype}      
\usepackage{subcaption}
\usepackage{amsmath,amsfonts,amssymb}
\usepackage{breqn}
\usepackage{tabularx}
\usepackage{multirow}
\usepackage{graphicx}
\usepackage{color}
\usepackage{tcolorbox}
\usepackage{CJK}
\usepackage{adjustbox}
\usepackage{xcolor}
\usepackage{colortbl}
\usepackage{multicol}
\usepackage{vwcol} 
\newsavebox{\algleft}
\newsavebox{\algright}
\usepackage{bm}
\usepackage{booktabs}
\usepackage{amsmath,stackengine}
\usepackage[linesnumbered,ruled,vlined]{algorithm2e}
\usepackage{times}
\usepackage{latexsym}
\usepackage{bbm}
\usepackage[T1]{fontenc}    
\usepackage{hyperref}       
\usepackage{url}            
\usepackage{booktabs}       
\usepackage{amsfonts}       
\usepackage{nicefrac}       
\usepackage{microtype}      
\usepackage{xcolor}         
\usepackage{amssymb}
\usepackage{pifont}
\usepackage{bbm}
\usepackage{breqn}
\usepackage{longtable}
\usepackage{algpseudocode}
\usepackage{enumitem}

\usepackage{microtype}
\usepackage{enumitem}
\usepackage{amsmath}

\title{NAACL2025 Tutorial: Adaptation of  Large Language Models}

\author{
Zixuan Ke ~~~~~~~~~~~~~ Yifei Ming ~~~~~~~~~~~~~  Shafiq Joty\\ 
Salesforce AI Research\\
\texttt{\{zixuan.ke,yifei.ming,sjoty\}@salesforce.com}\\  
Tutorial Page: \url{https://vincent950129.github.io/adapt-llm/}
}
\begin{document}
\maketitle





\section{Brief Description}

This tutorial on adaptation of Large Language Models (LLMs) is designed to address the growing demand for models that go beyond the static capabilities of generic LLMs by providing an overview of dynamic, domain-specific, and task-adaptive LLM adaptation techniques. While general LLMs have demonstrated strong generalization across a variety of tasks, they often struggle to perform well in specialized domains such as finance, healthcare, and code generation for underrepresented languages. Additionally, their static nature limits their ability to evolve with the changing world, and they are often extremely large in size, making them impractical and costly to deploy at scale. As a result, the adaptation of LLMs has drawn much attention since the birth of LLMs and is of core importance, both for industry, which focuses on serving its targeted users, and academia, which can greatly benefit from small but powerful LLMs. 




To our knowledge, there has been no similar tutorial on this topic in the past few years\footnote{\url{https://www.aclweb.org/adminwiki/index.php/Past_tutorials}}. To address this gap, this tutorial aims to provide an overview of the LLM adaptation techniques. We start with an introduction to LLM adaptation, from both the \textbf{data perspective}, which has been widely accepted as one of the most important ingredients in LLM training, and the \textbf{model perspective}, which focuses on the training strategies to adapt the LLMs.

In Session 2, we emphasize how the \textbf{evaluation metrics and benchmarks} are different from other techniques. There is a growing recognition that evaluating adapted LLMs should not focus solely on their performance within the target domains or tasks. While this remains important, an emerging consensus is that adaptation should not be at the cost of sacrificing what it has already learned. An ideal evaluation protocol should therefore assess not only specialized task performance but also the preservation of general knowledge. Furthermore, as we will present in this tutorial, adapting an LLM usually involves multiple training stages, where each stage contributes to the final outcome. This multi-stage process requires more comprehensive evaluation methods, as conventional metrics may not capture the cumulative impact of these stages. As such, researchers are increasingly focused on designing evaluation metrics and benchmarks that assess both the specialized knowledge gained and the general knowledge retained.

\begin{table}[t]
\centering
\resizebox{\columnwidth}{!}{
\begin{tabular}{l|l}
\toprule
 \textbf{Slot} & \textbf{Session} \\ \toprule
\multicolumn{2}{l}{\textit{Session 1: Introduction and Motivation}} \\
14:00 - 14:05 & Tutorial presenters introduction \\
14:05 - 14:10 & Challenges with general-purpose LLMs \\
14:10 - 14:20 & Overview and use cases of LLM adaptation \\
14:20 - 14:30 & Key concepts \\
14:30 - 14:40 & Research questions in data and model \\
\hline
\multicolumn{2}{l}{\textit{Session 2: Evaluation and benchmark }} \\
14:40 - 14:50 & General and specialized knowledge \\
14:50 - 15:00 & Evaluation metrics \\
\hline
\multicolumn{2}{l}{\textit{Session 3: Parametric Knowledge Adaptation}} \\
15:00 - 15:15 & Domain-adaptive Pre-training \\
15:15 - 15:30 & Instruction Tuning\\
15:30 - 16:00 & Coffee Break \\
16:00 - 16:15 & Preference Learning \\
16:15 - 16:30 & Model Editing \\
\hline
\multicolumn{2}{l}{\textit{Session 4: Semi-Parametric Knowledge Adaptation}} \\
16:30 - 16:45 & Retrieval-augmented Generation \\
16:45 - 17:00 & Agent-based Integration\\
\hline
17:00 - 17:30 & Summary, Discussion \& QA \\
\bottomrule
\end{tabular}
}
 \vspace{-4mm}
\caption{Example tutorial schedule. 
}
 \vspace{-8mm}
\label{tab.schedule}
\end{table}

After establishing the problems in the aforementioned sessions, we explore various \textbf{adaptation techniques}. When powerful general LLMs first emerged, many believed that prompt engineering alone, without any actual model training (i.e., updating the model’s parameters), was sufficient to adapt them for new tasks. However, recently we have seen more researchers begin finetuning model wights. This shift is understandable not only because training typically leads to better results on specialized tasks, but also because training has become more affordable due to the growing number of techniques that enable efficient training. For example, the training cost increased from \$900 for the original Transformer to over \$4 million to train the GPT-3 (davinci), but it has recently dropped to around \$0.8 million to train a GPT-3 on-par model like Phi3.5 ~\cite{standford2024aireport}. In light of this trend, this tutorial will focus on increasingly popular methods that adapt LLMs through training.

We categorize adaptation techniques into two main families. The first is \textbf{parametric knowledge adaptation}, which focuses on updating the parametric knowledge within LLMs, including methods like Domain-Adaptive Pre-Training (DAPT), Instructional Tuning (IT), and Preference Learning (PL) via human or model feedback. Additionally, we will discuss real-time adaptation techniques, including model editing, which allows LLMs to be updated dynamically in production environments. The second kind of adaptation is \textbf{semi-parametric knowledge adaptation}, where the goal is to update LLM parameters to better leverage external knowledge or tools (e.g., documents or functions) through techniques like retrieval-augmented generation (RAG) and agent-based systems.

\section{Detailed Outline}



This will be a \textbf{three-hour} tutorial devoted to the \textbf{cutting-edge topic} of the adaptation of LLMs. 
Table~\ref{tab.schedule} gives an overview. 


\setlist[enumerate,1]{left=0pt, itemsep=0em} 
\setlist[enumerate,2]{left=-14pt, itemsep=0em} 
\vspace{-2mm}

\begin{enumerate}
\item \textbf{Introduction and Motivation}
\vspace{-2mm}
\begin{enumerate}[label*=\arabic*.]
    \item \textbf{Challenges with General-purpose LLMs.} 
        While LLMs have demonstrated strong generalization capabilities, they are general and static. They often fall short in highly specialized domains or tasks. These LLMs struggle to adapt to evolving requirements and exhibit high computational overhead due to their fixed nature. This section will highlight key limitations of them, with detailed examples and statistics to illustrate why adaptation is necessary for handling domain-specific or task-specific demands.

    \item \textbf{Overview and Use Cases of LLM Adaptation.} 
    LLM Adaptation adapts a pre-trained LLM to specialized use cases. It offers practical solutions for domain-specific applications, such as in finance and healthcare, as well as task-specific applications like retrieval-augmented generation and code generation. Personalized systems also benefit from adaptive models, which can fine-tune responses based on individual user needs. This section will showcase real-world use cases that demonstrate the critical need for LLM adaptability to address specialized scenarios.

    \item \textbf{Key Concepts.} 
        Before exploring detailed methodologies, we will introduce foundational concepts essential to understanding adaptive LLMs. These include distinctions between domain vs. task adaptation, continual learning vs. LLM adaptation, Parameter-Efficient Fine-Tuning (PEFT), different stages of post-training (e.g., Domain-Adaptive Pre-Training and Supervised Fine-Tuning), long-context LLMs, retrieval-augmented generation (RAG), agentic approach. These concepts will set the stage for the following technical sections. 

    \item \textbf{Research Questions in Data and Model Perspective.} 
    In this tutorial, we aim to address several critical research questions. In data perspective, we focus on what gives a good data mixture and how to obtain high-quality data. In model perspective, we ask what constitutes an effective training recipe for adaptive LLMs, how can we properly evaluate adaptive models across different domains and tasks, what strategies can be used to prevent data leakage during post-training, and how can we scale up post-training efficiently. These questions will guide the subsequent discussions on evaluation and training.

\end{enumerate}
    \vspace{-3mm}

\item\textbf{Evaluation and Benchmarks.} Evaluating adaptive LLMs involves a multidimensional approach. We will cover methods for assessing generalization, domain- and task-specific performance, and real-time effectiveness. 
\vspace{-1mm}
    \begin{enumerate}[label*=\arabic*.]
    \item \textbf{General and Specialized Knowledge.} When adapting a model to specialize in specific knowledge, it is critical not to lose already learned general knowledge. A robust evaluation protocol should account for both. We will introduce methods to evaluate both types.
    \item \textbf{Evaluation Metrics
}. Numerous metrics have been discussed in the literature, such as perplexity, needle-in-haystack evaluations, and downstream task performance. We will compare and analyze the effectiveness of these metrics in evaluating adaptive LLMs.
    \end{enumerate}
\vspace{-1mm}

\item \textbf{Parametric Knowledge Adaptation.} Multiple stages are undertaken to refine and specialize their knowledge for specific tasks and domains.
\vspace{-5mm}

    \begin{enumerate}[label*=\arabic*.]
    \item \textbf{Domain-Adaptive Pre-Training (DAPT).} Discuss the data and objective of DAPT that learn domain-specific background knowledge using raw text with next token prediction. 
    \item \textbf{Instruction Tuning (IT).} Discuss the data and objective of IT, also known as Supervised Fine-Tuning (SFT). It can be full model fine-tuning or parameter-efficient fine-tuning, with focus on instruction following ability.
    \item \textbf{Preference Learning (PL)} Discuss the PL, where extra supervision on negative cases and delayed reward are introduced in this stage. The preference feedback can come from humans, AI models, or programmatic metrics. It can be done in online or offline (direct preference optimization; DPO) manner.
    \item \textbf{Model Editing.} Discuss the techniques that allow for targeted updates in real-time, improving adaptability in production environments without the need for expensive retraining. 
    \end{enumerate}



\item \textbf{Semi-parametric Knowledge Integration}
    \begin{enumerate}[label*=\arabic*.]
    \item \textbf{Retrieval-augmented Generation (RAG)} How RAG allows models to dynamically retrieve external knowledge, and significantly enhance performance on specialized tasks or when up-to-date information is needed.
    \item \textbf{Agent-based Integration.} This focuses more on the agentic aspect, highlighting the model as an agent that performs multi-step reasoning and interacts with external APIs and dynamic environments. It draws attention to the system's autonomous decision-making capacity and its ability to operate as an agent in complex environments.
    \end{enumerate}



\item \textbf{Summary, Discussion \& QA} 
\end{enumerate}

\section{Proportion of Other Researchers' Work, Reading list and Prerequisite}



LLM adaptation involved almost all domains and tasks, making it become the main focus of industry and academic research. Our tutorial is designed to provide a comprehensive overview of this broad and rapidly evolving field, ensuring that the content reflects a wide range of techniques and contributions. As a result, the tutorial naturally includes a substantial amount of work from other researchers.

The prerequisite includes familiarity with basic knowledge of
machine learning, NLP and LLM. Knowledge of continual learning is a plus.  Here are a
few papers that lay a foundation for this area and we encourage attendees to read the papers before the tutorial to familiarize themselves with foundational concepts and approaches:
\vspace{-3mm}

\begin{itemize}[leftmargin=*]
    \item Continual pre-training of language models \cite{ke2023continual}
    \vspace{-3mm}

    \item Locating and editing factual associations in GPT \cite{meng2023locatingeditingfactualassociations}
    \vspace{-3mm}

    \item Retrieval augmented language model pre-training \cite{guu2020retrieval}
    \vspace{-3mm}


\end{itemize}

Below, we provide a detailed reference to the works we plan to cover. While we will give pointers to dozens
of relevant papers over the course of the tutorial, we plan to cover around 7-8 research papers in close detail. Only 1-2 of the ``deep dive'' papers will come from the presenter team.





\textbf{Continual Learning} learns a sequence of tasks sequentially without forgetting~\citep{chen2018lifelong,mccloskey1989catastrophic,van2019three,mai2022online,DBLP:conf/nips/AljundiLGB19,ke2022survey}, which is closely related to LLM adaptation. Typical approaches include \emph{regularization-based methods} that regularize parameter updates to preserve important parameters~\citep{Kirkpatrick2017overcoming,Seff2017continual}; \emph{modular-based methods} that dynamically modify the architecture~\citep{Serra2018overcoming,wortsman2020supermasks}; and \emph{replay-based method} that recall previous experiences~\citep{Rebuffi2017,wang2020efficient}.

\textbf{Parametric Knowledge Adaptation} adapts LLMs by updating its parametric knowledge. It has been widely adopted across board domains, such as code~\citep{nijkamp2022codegen}, medical~\citep{luo2023biomedgpt}, law~\citep{colombo2024saullm}, mathematics~\citep{azerbayev2023llemma}, multi-lingual~\citep{chen2024effectiveefficientcontinualpretraining} and finance~\citep{ke2025demystifying,xie2023pixiulargelanguagemodel,Palmyra-Fin-70B-32k} and tasks such as function calling~\citep{zhang2024xlam}. Some focus on domain-specific or task-specific data curation~\citep{yang2024synthetic}, mixture ratio~\citep{que2024dcptlawdomainspecificcontinual}, data-efficiency~\citep{xie2023efficientcontinualpretrainingbuilding}, hyper-parameters~\citep{parmar2024reuse} or training recipe \cite{jiang2024instruction,yu2024lionsempiricallyoptimizedapproach,gao2024trainlongcontextlanguagemodels}.

\textbf{Semi-Parametric Knowledge Adaptation} augments LLMs with relevant information retrieved from various knowledge sources, i.e., RAG or external functions, i.e., agent-based intergation.

\textbf{Retrieval-Augmented Generation (RAG)} is proven effective across numerous NLP tasks, including language modeling \cite{borgeaud2022improving,Khandelwal2020Generalization,shi2023replug}, question answering \cite{bridging_retriever_llm_ke2024,lewis2020retrieval,izacard2022few,de2023glimmer,de2023pre,shi2023replug,guu2020retrieval,izacard2020leveraging,xu2023recomp}, fact versification \cite{lewis2020retrieval} and text generation \cite{lewis2020retrieval}. Specifically, RAG utilizes input as a query and comprises two main components: a \emph{retriever} retrieves a set of items from a side corpus,
 and a \emph{LLM} incorporates the retrieved items, as additional information in the input context, and makes final predictions. Depending on which components are subject to updates, typical approaches can be categorized into three families: \emph{adapting retrievers and LLMs jointly}, which is the most widely used setting of RAG \cite{izacard2022few,Khandelwal2020Generalization,wu2022memorizing,guu2020retrieval,lewis2020retrieval}; \emph{adapting LLMs only}, which avoid the costly updates of retrievers and document index \cite{izacard2020leveraging,de2023pre,de2023glimmer}; and \emph{adapting retrievers only}, which compatible with black-box LLMs~\cite{shi2023replug,xu2023recomp}.

\textbf{Model Editing} aims to adapt LLM in real-time without the need for expensive training or even without training \cite{zhang2024comprehensivestudyknowledgeediting,wang2024knowledgeeditinglargelanguage}. Typical approaches include \emph{locate-and-edit} that first locate the position of the knowledge in LLMs and edit the KV cache \cite{meng2023locatingeditingfactualassociations,meng2023masseditingmemorytransformer,yang2024butterflyeffectmodelediting,gupta2024rebuildingromeresolving,gupta2024unifiedframeworkmodelediting}, \emph{meta-learning} that employ hyper-network to learn the editing \cite{tan2024massiveeditinglargelanguage,decao2021editingfactualknowledgelanguage,mitchell2022fastmodeleditingscale} and \emph{memory-based} that use memory elements to store and manipulate information during
editing \cite{zhu2020modifyingmemoriestransformermodels,ni2024forgettinglearningutilizingparametric,zheng2023editfactualknowledgeincontext}. As of now, editing is still in its infancy and suffers from serious forgetting, in particular when editing a sequence of samples.

\textbf{Agent-based Integration} augments the LLMs with external functions. Typical approaches include \emph{fine-tuning the LLMs} for better function calling capacities \cite{qin2023toolllmfacilitatinglargelanguage,chen2023fireactlanguageagentfinetuning,patil2023gorillalargelanguagemodel,zeng2023agenttuningenablinggeneralizedagent,zhang2024agentohanadesignunifieddata,yin2024agentlumosunifiedmodular,chen2024agentflandesigningdatamethods,li-etal-2023-api,tang2023toolalpacageneralizedtoollearning}; and \emph{prompting techniques} \cite{wei2023chainofthoughtpromptingelicitsreasoning,yao2022react,shinn2023reflexion}


\vspace{-3mm}





\section{Target Audience and Estimated Size}

Researchers, graduate students, and practitioners who are interested in NLP and machine learning. The tutorial will is particularly beneficial to people who intend to develop machine learning and NLP techniques and applications that can adapt the LLM to specialized domains or tasks while preserving the original learned knowledge in the LLM. 
This tutorial has not been given before. Due to the increasing popularity of LLM and its adpatation, we estimate that there will be 100 to 150 people attending the tutorial.

\vspace{-2mm}
\section{Diversity Consideration}

The topic of LLM adaptation will be inclusive for a broad range of communities. It will be of interest to industrial practitioners as they aim to deploy the best adaptive models for their targeted customers. Additionally, academic researchers will find value in this topic, as it essentially addresses a continual learning problem, enabling LLMs to adapt to new environments without forgetting—an essential step toward advancing Artificial General Intelligence. 
This topic also makes it easier for users to engage with LLMs, as it involves techniques to train smaller models that can be deployed efficiently. It naturally spans a wide range of domains and tasks, making it relevant to participants interested in various specific areas.

Our presenters include both junior and senior researchers from industry and academia, and we also bring geographical diversity, with instructors from both North America and Asia. This diverse group of instructors will help attract a wide audience and encourage participation from various backgrounds. 

We will make
our tutorial materials digitally accessible to all participants. All materials will be publicly available before the tutorial. We will provide slides and discussion questions before the tutorial to the audience. During the tutorial sessions, we will work with student volunteers to encourage open dialogue
and promote active listening, allowing participants to share their thoughts and experiences without fear
of judgment. After the tutorial, we will actively collect feedback to identify areas for improvement related to diversity and inclusion and share it with future tutorial presenters. Our presenter team will share our tutorial with a worldwide audience by promoting it on social media, and to diverse research communities. We will work with ACL/NAACL/EMNLP D\&I teams, and consult resources such as the BIG directory to diversify our
audience participation.

\vspace{-2mm}
\section{Ethics Statement}


The tutorial will present general problems and algorithms related to LLM adaptation in NLP. All evaluation datasets are public domain benchmarks. Thus, the tutorial does not have any ethical issues.

\section{Presenter Biography}


\textbf{Zixuan Ke} is a research scientist at Salesforce AI Research, where he works with Dr. Shafiq Joty on continual pre-training of LLMs. Prior to that, he earned his Ph.D. at the University of Illinois, Chicago, where he worked with Dr.~Bing Liu on continual learning. His research studies how to adapt the foundation models, particularly LLMs, for an ever-changing world characterized by emerging domains, events, topics, or information.
He has published 25+ papers in top-tier conferences such as ICLR, ICML, NeurIPS, NAACL, ACL, and EMNLP. 
He has served as an Area Chair and PC member for prestigious conferences such as ICLR, NeurIPS, ACL, EMNLP, and as a reviewer for leading journals such as Neural Networks, Neurocomputing, Artificial Intelligence, TKDE, and TPAMI, since 2021. He has given an invited short PhD course on continual learning at Aalborg University in the summer of 2022 and has also given several oral presentations at leading conferences. Further information about him, including past talks and slides, can be found at \url{https://vincent950129.github.io/}.

\textbf{Yifei Ming} is a Research Scientist at Salesforce AI Research. He received his Ph.D. in Computer Science from the University of Wisconsin-Madison in 2024. His research spans both empirical and theoretical studies, with a focus on understanding, benchmarking, and developing novel algorithms towards trustworthy foundation models (LLM and VLM). He has a prolific publication record with 20+ papers in top-tier AI conferences and journals. He co-organized several workshops and has served on the program committees of leading machine learning conferences and journals since 2021, including NeurIPS, ICLR, ICML, AAAI, EMNLP, IJCV, TPAMI, and NN. Further information can be found at \url{https://alvinmingsf.github.io/}.

\textbf{Shafiq Joty} is a research director at Salesforce Research and an Associate Professor at the Computer Science Department of the Nanyang Technological University. His research has contributed to over 30+ patents and 150+ papers in top-tier NLP and ML conferences and journals. He has served as a Program Chair of SIGDIAL'23, a member of the best paper award committees for ICLR'23 and NAACL'22, and in the capacity of a (senior) area chair for many leading NLP and ML conferences (e.g. NeurIPS, EMNLP, and ACL). He previously gave tutorials at EMNLP'23, IEEEVis'22, ACL'19,  COLING'18 and ICDM'18. Further information about him, including past talks and slides, can be found at \url{https://raihanjoty.github.io/}.

\bibliography{anthology,custom}
\bibliographystyle{acl_natbib}




\end{document}